\documentclass[sigconf]{acmart}
\AtBeginDocument{%
  }

\usepackage{amsmath,amsfonts,bm}

\def\eqref#1{equation~\ref{#1}}

\def\1{\bm{1}}

\def\vh{{\bm{h}}}

\DeclareMathAlphabet{\mathsfit}{\encodingdefault}{\sfdefault}{m}{sl}
\SetMathAlphabet{\mathsfit}{bold}{\encodingdefault}{\sfdefault}{bx}{n}

\def\gB{{\mathcal{B}}}

\def\gE{{\mathcal{E}}}

\def\gG{{\mathcal{G}}}

\def\gL{{\mathcal{L}}}

\def\gT{{\mathcal{T}}}

\def\gV{{\mathcal{V}}}

\usepackage{url}
\usepackage{booktabs}
\usepackage{graphicx}
\usepackage{algorithm}
\usepackage{algpseudocode}
\usepackage{amsmath}
\usepackage{amsfonts}
\usepackage{multirow}
\usepackage{wrapfig}
\usepackage{caption}
\usepackage{subcaption}
\usepackage{pifont}
\usepackage{cleveref}
\usepackage[textsize=tiny, color=green!30]{todonotes}
\usepackage{enumitem}
\setitemize{topsep=0pt, partopsep=0pt, leftmargin=12pt, itemsep=2pt}
\setenumerate{topsep=0pt, partopsep=0pt, leftmargin=12pt, itemsep=2pt}

\newcommand{\tinycolorbox}[2]{\tikz[baseline=(a.base),inner sep=0pt]\node[fill=#1](a){#2};}

\def\cmark{{\color{green}\ding{51}}}%
\def\xmark{{\color{red}\ding{55}}}%
\setuptodonotes{inline, color=yellow!40!white}

\def\method{\textbf{SemStruct}}

\copyrightyear{2026}
\acmYear{2026}
\setcopyright{cc}
\setcctype{by}
\acmConference[KDD '26]{Proceedings of the 32nd ACM SIGKDD Conference on Knowledge Discovery and Data Mining V.2}{August 09--13, 2026}{Jeju Island, Republic of Korea}
\acmBooktitle{Proceedings of the 32nd ACM SIGKDD Conference on Knowledge Discovery and Data Mining V.2 (KDD '26), August 09--13, 2026, Jeju Island, Republic of Korea}
\acmDOI{10.1145/3770855.3817963}
\acmISBN{979-8-4007-2259-2/2026/08}

\begin{document}

\title{SemStruct: Contextualizing Semantic Embeddings with Structural Information for Schema Matching}


\author{Inwon Kang}
\email{kangi@rpi.edu}
\orcid{0000-0001-8912-287X}
\affiliation{%
	\institution{Rensselaer Polytechnic Institute}
	\city{Troy}
	\state{New York}
	\country{USA}
}

\author{Kavitha Srinivas}
\email{kavitha.srinivas@ibm.com}
\orcid{0000-0003-4610-967X}
\affiliation{%
	\institution{IBM Research}
	\city{Yorktown Heights}
	\state{New York}
	\country{USA}
}

\author{Nandana Mihindukulasooriya}
\email{nandana@ibm.com}
\orcid{0000-0003-1707-4842}
\affiliation{%
	\institution{IBM Research}
	\city{New York City}
	\state{New York}
	\country{USA}
}

\author{Sola Shirai}
\email{solashirai@ibm.com}
\orcid{0000-0001-6913-3598}
\affiliation{%
	\institution{IBM Research}
	\city{Yorktown Heights}
	\state{New York}
	\country{USA}
}

\author{Parikshit Ram}
\email{parikshit.ram@ibm.com}
\orcid{0000-0002-9456-029X}
\affiliation{%
	\institution{IBM Research}
	\city{Yorktown Heights}
	\state{New York}
	\country{USA}
}

\author{Horst Samulowitz}
\email{samulowitz@us.ibm.com}
\orcid{0000-0002-6780-3217}
\affiliation{%
	\institution{IBM Research}
	\city{Yorktown Heights}
	\state{New York}
	\country{USA}
}

\author{Oshani Seneviratne}
\email{senevo@rpi.edu}
\orcid{0000-0001-8518-917X}
\affiliation{%
	\institution{Rensselaer Polytechnic Institute}
	\city{Troy}
	\state{New York}
	\country{USA}
}

\renewcommand{\shortauthors}{Kang et al.}

\begin{abstract}
	Schema matching is a fundamental step in integrating heterogeneous data sources. While Pre-trained Language Models (PLMs) have revolutionized this task by capturing linguistic semantics, they typically process tabular data as serialized text sequences of standalone column descriptions. This serialization discards critical structural information -- specifically, the row-level co-occurrences, i.e. the relational context -- forcing models to rely solely on column header semantics or standalone distributions. To bridge this gap, we propose $\textbf{SemStruct}$, a framework that joins the semantic power of frozen PLMs with the structural inductive bias of Graph Neural Networks (GNNs). We model the table as a heterogeneous graph where columns and values are nodes connected by rows, allowing the GNN to propagate disambiguating context across the structure. Unlike other state-of-the-art methods that require proprietary LLM access and fine-tuning of language models, SemStruct keeps the language model frozen and trains only a lightweight structural encoder. Extensive experiments on the Valentine and SOTAB-SM benchmarks demonstrate that SemStruct achieves state-of-the-art performance, outperforming fully fine-tuned baselines on complex, semantically joinable datasets. Furthermore, our ablation studies reveal that row representations serve primarily as topological conduits rather than semantic entities, validating the necessity of explicit structural modeling in schema matching.
\end{abstract}



\received[accepted]{15 May 2026}

\maketitle

\section{Introduction}

Automated table parsing and schema matching are critical for integrating heterogeneous data sources. While Pre-trained Language Models (PLMs) and dense retrieval methods have improved semantic matching, these methods typically treat tabular data as serialized text sequences. This approach effectively linearizes the 2D table structure, discarding the explicit relational dependencies that exist between the columns of a table. As a result, these methods will fail to capture this structural information that may be valuable for disambiguating between columns with similar or ambiguous semantics.

\begin{figure}[h]
	\centering
	\includegraphics[width=\linewidth]{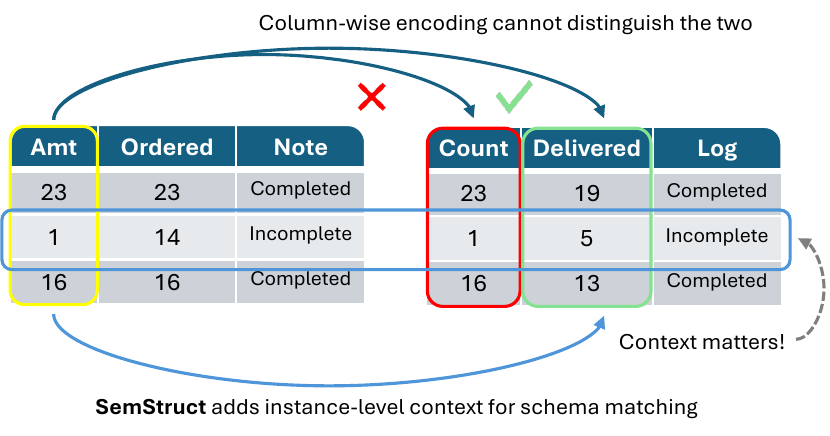}
	\caption{Example of a scenario where row-context matters. The content of the row can help disambiguate incomplete information.}
	\label{fig:row-instance-example}
\end{figure}

Existing approaches generally fall into two categories: heuristic methods relying on string similarity, and semantic methods using PLMs to embed column headers and values. While PLMs capture linguistic semantics, they often fail to leverage the structural context defined by the rows. For instance, consider two tables containing numerical columns with ambiguous names such as \textit{value} or \textit{amt}. Without additional context, it is challenging to determine the exact semantics of such columns. However, the ambiguity is resolved by analyzing \textit{row-level value interactions} of these columns with other columns. Consider the example illustrated in~\Cref{fig:row-instance-example}, where we need to match the left table's \textit{``Amt''} column to one of the right table's columns. Looking at each column separately, one may think that \textit{``Amt''} corresponds to \textit{``Count''}. However, when looking at the accompanying values, in this case, the \textit{``note''} or \textit{``log''} column, we can infer that \textit{``Amt''} refers to the \textit{amount of items delivered}, while \textit{``Count''} refers to the \textit{number of items ordered}. This example illustrates that row-level interactions provide crucial context for disambiguating column semantics, which column-only serialization approaches overlook.

To address this, we propose~\method{}, a framework that combines frozen sentence encoders with Graph Neural Networks (GNNs). Unlike state-of-the-art methods like ISResMat~\cite{duSituNeuralRelational2024} or Magneto~\cite{liuMagnetoCombiningSmall2025}, which solely rely on column-level embeddings and ignore row-level interactions,~\method{} joins structural reasoning on top of the semantic priors of the language model by introducing a graph layer to model the \textit{interactions}. We construct a heterogeneous graph where columns, rows and column values are nodes, where each row node serves as a bridge connecting the different column value nodes present in that row. This allows the model to propagate context explicitly: a value node can update the representation of its parent column based on the other values present in the same row. By keeping the LM frozen and training only the lightweight GNN, we achieve high accuracy without the prohibitive cost of fine-tuning Transformers on large data lakes, as well as retaining the semantic priors of the LM.

Our contributions are as follows:
\begin{itemize}
	\item \textbf{Graph-Based Inductive Bias:} We propose \method{}, which models schema matching as a message-passing problem on a heterogeneous graph. This explicitly captures the row-level value co-occurrence that serialized methods neglect.
	\item \textbf{Structural Insights:} We identify that row representations act as ``structural conduits'' rather than semantic entities in our method. Our ablation studies show that initializing row nodes with zero vectors outperforms semantic aggregation, suggesting that the row nodes' primary role is topological rather than semantic.
	\item \textbf{SOTA Performance:} We evaluate \method{} on the Valentine and SOTAB benchmark, showing that it outperforms baselines on complex, semantically-joinable datasets without requiring proprietary LLM access or fine-tuning.
\end{itemize}

\section{Related Work}

Existing schema matching literature is broadly categorized into heuristic-based methods, language model-based approaches, and graph-based representation learning.

\subsection{Heuristic and Rule-Based Matching}

Traditional schema matching relies heavily on linguistic similarity and structural constraints. \citet{rahm2001survey} established the foundational taxonomy, distinguishing between element-level and structure-level matchers. These systems typically employ string similarity metrics (e.g., Jaccard, Levenshtein) or thesauri to align attributes~\cite{bernstein2011generic}. While effective for simple naming conventions, these methods lack semantic generalization and fail when attribute names are opaque or ambiguous. \citet{koutrasValentineEvaluatingMatching2021} formalized the evaluation of such methods with \textit{Valentine}, a benchmark suite categorizing datasets into scenarios such as unionable and semantically-joinable tables, highlighting the performance degradation of heuristic baselines on semantically complex data. While Valentine offers a comprehensive set of evaluations for different matching scenarios, its synthetically generated matching pairs are often \textit{easy}, as the column header values can still provide significant semantic clues. In this regard, \citet{rengWDCSchemaMatching} offer a more realistic benchmark -- the Web Data Commons Schema Matching Benchmark (WDC-SMB) -- composed of tables derived from the Web. In particular, the Schema.org Table Annotation Benchmark (SOTAB)~\cite{koriniSOTABWDCSchemaorg2022} provides challenging table matches based on Schema.org properties, where column headers are replaced with simple numbers, making it difficult to rely on header semantics alone.

\subsection{Language Models for Table Representation}

Pre-trained Language Models (PLMs) have been adapted to encode tabular data by serializing table substructures. Models such as TaBERT~\cite{yinTaBERTPretrainingJoint2020}, TAPAS~\cite{herzig2020tapas}, and TAPEX~\cite{liuTAPEXTablePretraining2021} are pre-trained on large corpora to optimize tasks like semantic parsing and question answering. However, these architectures generally require task-specific fine-tuning and architectural modifications to handle 2D table structures.

In the context of schema matching and entity resolution, deep learning approaches leverage PLMs to capture semantic dependencies. Ditto~\cite{li2020deep} fine-tunes BERT with domain-specific data augmentation for entity matching. SMAT~\cite{zhang2019heterogeneous} employs a Siamese network combining BERT embeddings with BiLSTMs. More recently, ISResMat~\cite{duSituNeuralRelational2024} proposed a framework that fine-tunes a PLM directly on target table pairs. It utilizes pairwise sampling to generate column fragments and optimizes a \textit{Meta-Matching Loss} alongside an \textit{Agent-Delegating Loss} to align column representations without external training data. Conversely, methods like Starmie~\cite{fan2023semantics} focus on table union search in data lakes using contrastive learning to retrieve unionable tables, prioritizing retrieval efficiency over granular schema alignment.

While fine-tuning PLMs yields high semantic fidelity, it is computationally intensive and susceptible to overfitting on small schema matching datasets. Alternatively, Magneto~\cite{liuMagnetoCombiningSmall2025} utilizes frozen or lightly-tuned small language models within a retrieve-and-rerank pipeline, reducing computational overhead but potentially limiting representational depth compared to fully trained encoders. However, it is worth noting that both ISResMat and Magneto are limited to encoding column-level semantics -- they do not explicitly model row-level co-occurrences that could provide critical relational context for disambiguating column meanings.

\subsection{Graph Representation for Relational Data}

Graph Neural Networks (GNNs) provide a natural inductive bias for relational data by explicitly modeling structural dependencies. Heterogeneous graph architectures, such as R-GCN~\cite{schlichtkrull2018modeling} and HGT~\cite{huHeterogeneousGraphTransformer2020}, handle diverse node types (rows, columns, values), making them suitable for tabular representation. \citet{fey2023relational} theoretically demonstrated that GNN message passing aligns with relational algebra operations like JOIN and AGGREGATE.

Specific to schema matching, REMA~\cite{koutrasREMAGraphEmbeddingsbased2020} constructs a heterogeneous graph where nodes represent schema elements and data values. It generates embeddings via random walks (DeepWalk-style) rather than deep semantic encoding. While REMA effectively captures structural similarity, it trains embeddings from scratch, thereby failing to leverage the pre-trained semantic knowledge encapsulated in LLMs. Similarly, SANTOS~\cite{khatiwada2023santos} utilizes knowledge graphs for table union search, modeling relationships between columns to infer semantic intent.

Current state-of-the-art approaches typically force a trade-off: PLM-based methods (e.g., ISResMat) excel at semantics but often require task-specific fine-tuning that overfits to training schemas, or neglect global structural signals, while GNN-based methods (e.g., REMA) capture structure but lack pre-trained semantic understanding. Our work addresses this gap by combining the semantic richness of frozen PLMs with the structural reasoning of trained GNNs.

\subsection{Schema Matching}
We adopt the problem definition provided by~\citet{koutrasValentineEvaluatingMatching2021}. Let $T_S = \{c_1^S, \dots, c_n^S\}$ be a source table and $T_T = \{c_1^T, \dots, c_m^T\}$ be a target table, where $c_i$ denotes a column. \textit{Schema matching} is the task of finding a set of pairs $M = \{(c_i^S, c_j^T)\}$ such that the columns in each pair are semantically equivalent or joinable.


\subsection{Evaluation Metrics}

We evaluate performance using two standard metrics: Mean Reciprocal Rank (MRR) and Recall@GT (Recall at Ground Truth).
\begin{itemize}
	\item \textbf{MRR} measures the average reciprocal rank of the correct match in the predicted list, providing insight into the ranking quality of the model. This metric tells us \textit{how correct the top matches are for each column}. 
	\item \textbf{Recall@GT} calculates the proportion of ground truth matches successfully retrieved within the top-$k$ predictions, where $k$ is the number of true matches for a given source column. This metric reflects the \textit{model's ability to capture all relevant matches} without aggressive filtering.
\end{itemize}

\section{Methodology}
The~\method{} framework consists of three main components: (1)~Graph Construction from tabular data, (2)~Semantic Initialization using frozen PLMs, and (3)~Structural Refinement via GNNs.

\begin{figure*}
	\centering
	\includegraphics[width=0.95\linewidth]{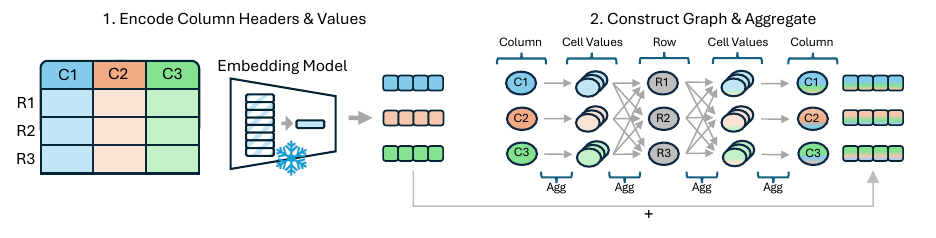}
	\caption{Visual description of the~\method{} inference on a single table. 1. The column node and cell node features are initialized using a frozen PLM. 2. The node features are updated via a GNN over the Row-Column-Value graph. The final column embeddings are obtained via a residual connection between the semantic and structural embeddings.}
	\label{fig:method}
\end{figure*}

\subsection{Tabular Graph Construction}
To capture the structural dependencies within a table $T$, we first convert the table into a heterogeneous graph $\gG = (\gV, \gE)$, similar to the graph seen in~\cite{wuOpenWorldFeatureExtrapolation2021,koutrasREMAGraphEmbeddingsbased2020}. The node set $\mathcal{V}$ consists of three types of nodes:
(1) \textbf{Column Nodes ($\gV_C$):} Representing the attributes the table describes. This node is the primary focus that we wish to enrich for schema matching;
(2) \textbf{Row Nodes ($\gV_R$):} Representing the individual data points (records) in the table; and
(3) \textbf{Cell (value) Nodes ($\gV_V$):} Representing the unique values present in \textit{each column.}

For numerical columns, we apply a quantile-based discretization strategy with $k=8$ to convert continuous values into discrete tokens, treating it as a categorical column of ranges, ensuring that the number of value nodes does not explode for numerical columns.

The edge set $\mathcal{E}$ defines the interactions between these nodes:
\begin{itemize}
	\item Edges between Column nodes and Cell nodes $(c_i, v_{ij})$ exist if value $v_{ij}$ appears in column $c_i$.
	\item Edges between Row nodes and Cell nodes $(r_j, v_{ij})$ exist if value $v_{ij}$ appears in row $r_j$.
\end{itemize}

This construction allows information to flow from columns to values, values to rows, and back to other values and columns, explicitly modeling the row-level co-occurrence (context). We use bidirectional edges to facilitate the propagation of information in both directions.

\subsection{Node Feature Initialization}
We leverage a PLM to initialize the column and value node features. The column and value nodes are initialized by encoding their respective textual representations derived from serialization. However, unlike columns and values, it is not immediately clear how the row can be represented in the same space -- a row represents a \textit{collection} of different values/columns. Thus, we explore several strategies for initializing the row nodes: zero-init, random-init, and semantic-init (mean-pooling connected value nodes).

\subsubsection{Node Merging} In practice, we find that the graph can become quite sparse, especially when there are columns with high cardinality values. This would lead to inefficient message passing in the GNN, as it will struggle to extract interaction information from rarely connected nodes. To mitigate this, we use a node merging strategy
~\cite{koutrasREMAGraphEmbeddingsbased2020}, where we merge cell node pairs that have high cosine similarity above some threshold $\tau$. This effectively reduces the number of unique cell nodes while preserving the semantic relationships between similar values. We set $\tau=0.9$ as the default parameter in our experiments.

\subsubsection{Serialization}
The choice of serialization has proven to be a crucial factor in the performance of LMs on tabular tasks~\cite{liuMagnetoCombiningSmall2025,suiTableMeetsLLM2024}. We also experiment with different strategies for serializing the column headers before they are embedded.~\Cref{tab:serialization_desc} describes the column node serialization strategies considered in our work.

For the value nodes, we use a simple serialization format of \texttt{``column: \{\}, value: \{\}''} to provide both the column context and the actual value to the PLM embedder.

\begin{table}
	\centering
	{\small
		\begin{tabular}[c]{ll}
			\toprule
			\textbf{Strategy}                 & \textbf{Description}                                                                                                          \\ \midrule
			\texttt{header}                   & Name only. \tinycolorbox{blue!30}{``\{\texttt{COL\_NAME}\}''}                                                                 \\
			\midrule
			\multirow{4}{*}{\texttt{verbose}} & Name + statistical summary if numerical, list of all unique                                                                   \\
			                                  & values if categorical. \tinycolorbox{blue!30}{ \texttt{``Column: \{COL\_NAME\}, Min: \{\},}}                                  \\
			                                  & \tinycolorbox{blue!30}{\texttt{Max: \{\}, Mean: \{\}, \ldots''}} or \tinycolorbox{blue!30}{``\texttt{Column: \{COL\_NAME\},}} \\
			                                  & \tinycolorbox{blue!30}{\texttt{Unique Values: \{\}, \ldots''}}                                                                \\
			\midrule
			\multirow{5}{*}{\texttt{minmax}}  & Simpler version of \texttt{verbose}. Sample $k$ min/max values for                                                            \\
			                                  & that column, separated by $\ldots$ to indicate that we are listing                                                            \\
			                                  & the tail and head of the distribution. \tinycolorbox{blue!30}{``\texttt{Column: \{COL\_NAME\},}}                              \\
			                                  & \tinycolorbox{blue!30}{\texttt{\{Min\_1\}, \ldots \{Min\_k\} \ldots, \{Max\_k\}, \ldots \{Max\_1\}''}}.                       \\
			                                  & If categorical, list all unique values as in \texttt{verbose}.                                                                \\
			\midrule
			\multirow{3}{*}{\texttt{random}}  & Randomly sample $k$ values from the column.                                                                                   \\
			                                  & \tinycolorbox{blue!30}{``\texttt{Column: \{COL\_NAME\}, \{Val\_1\}, \{Val\_2\}, \ldots,}}                                     \\
			                                  & \tinycolorbox{blue!30}{\texttt{\{Val\_k\}''}}                                                                                 \\
			\bottomrule
		\end{tabular}
	}
	\caption{
		Description of different column header serialization strategies used for initializing column node features.
	}
	\label{tab:serialization_desc}
\end{table}

\subsection{Structural Refinement via GNN}
Now that we have constructed the tabular graph and initialized the node features, we need a way to \textit{refine} the column embeddings based on the structural context provided by the rows and cell values. To achieve this, we employ a Graph Neural Network (GNN) to update the node embeddings based on the graph structure (neighbors). The GNN serves to propagate ambiguity-resolving context from the `row' nodes (which represent tuple co-occurrence) back to the column nodes. In particular, we use a multi-layer GNN architecture. At layer $L$, the node embeddings are updated as:

\begin{equation}
	\vh_v^{(L)} = \text{GNN}\left(\vh_v^{(L-1)}, \{\vh_u^{(L-1)} | u \in \mathcal{N}(v)\}\right),
\end{equation}

where $\mathcal{N}(v)$ denotes the neighbors of node $v$. We experiment with different GNN variants, such as GraphSAGE~\cite{hamiltonInductiveRepresentationLearning2017}, Graph Attention Networks (GAT)
~\cite{velickovicGraphAttentionNetworks2018}, and Heterogeneous Graph Transformer (HGT)~\cite{huHeterogeneousGraphTransformer2020} to find the most effective aggregation function for our tabular graph.
Once we have obtained the final node embeddings $\vh_v^{(L)}$ after $L$ layers of GNN, we extract the refined column embeddings $\vh_c^{(L)}$ and add a residual connection to the initial semantic embeddings:
\begin{equation}
	\vh_c^{final} = \vh_c^{(0)} + \text{MLP}(\vh_c^{(L)}),
\end{equation}
where MLP is a simple 2-layer feedforward network that projects the GNN output. We employ a residual connection to preserve the strong semantic priors of the pre-trained encoder, forcing the GNN to learn a contextual delta rather than reconstructing semantics from scratch.

\subsection{Self-Supervised Training}
To train the GNN without requiring labeled schema matches across real-world datasets, we adopt a self-supervised approach inspired by Valentine~\cite{koutrasValentineEvaluatingMatching2021} and Magneto~\cite{liuMagnetoCombiningSmall2025}. That is, we generate synthetic views of the \textit{source} tables of the Valentine benchmark and train the model to match columns between the synthetic views that originated from the same source table. Specifically, given a table $T$, we generate a set of augmented views $\hat{\gT} = \{\hat{T}^1, \hat{T}^2 \ldots \}$ by randomly sampling rows and columns. We further augment each view by applying column-header and value level noise (typos, formatting changes, acronyms). Since these views are derived from the original table, we have access to the ground truth column mappings between $T$ and each $\hat{T}^i$. It is important to note that we only use the \textit{source} tables from the Valentine benchmark -- the evaluation ground truth pair (the \textit{target} tables) are never seen during training to ensure that there is no data leakage.

\subsubsection{Triplet Loss}
Similar to Magneto~\cite{liuMagnetoCombiningSmall2025}, we also opt to use a triplet loss for training our GNN module. Intuitively, the goal of triplet loss is to bring the embeddings of corresponding columns (positives) closer together while pushing apart the embeddings of non-corresponding columns (negatives). Given a batch of tables and their ground truth column mappings, we construct a batch containing multiple tables and their generated views. Let $\gB$ be the batch, $c_i$ an anchor column, $p_{c_i}$ the set of positive columns that correspond to $c_i$ in the augmented views, and $n_{c_i}$ the set of negative columns that do not correspond to $c_i$. For each anchor column $c_i$, we minimize the following objective:

\begin{equation}
	\gL = \sum_{c_i \in \gB} \left[
		\max_{c_p \in p_{c_i}}(\text{dist}(\vh_{c_i}, \vh_{c_p}))
		- \min_{c_n \in n_{c_i}}(\text{dist}(\vh_{c_i}, \vh_{c_n}))
		+ \alpha \right],
\end{equation}

where $\alpha$ denotes the \textit{margin} of the triplet loss and $\text{dist}(\cdot, \cdot)$ is a distance metric. We use cosine distance in our experiments and set the margin $\alpha=0.2$. Similar to Magneto, we also use hard positive (positive sample with the highest distance) and negative (negative sample with the lowest distance) mining within the batch to maximize the training signal.

\subsubsection{Batch Construction}
One challenge with triplet loss is the difficulty of mining the positive/negative samples effectively during training. We wish to maximize the training signal in each batch by ensuring that there is a sufficient number of positive pairs -- negative pairs are abundant since any non-matching pair is negative. Thus, we construct each batch such that every table has at least one positive match. For Valentine, since each fabricated table is generated from a source table, we can group the fabricated tables by their source table during batching.

\section{Experiments}
\label{sec:experiments}

\begin{table}[tp!]
	\centering
	{\small
		\begin{tabular}{lrrrrrr}
			\toprule
			Dataset  & \#Tasks & \#Rows  & \#Cols & Num/Cat & Cat. Card. \\
			\midrule
			ChEMBL   & 180     & 10500.0 & 17.4   & 0.5/0.5 & 1835.4     \\
			OpenData & 180     & 16277.7 & 38.2   & 0.6/0.4 & 141.9      \\
			TPC-DI   & 180     & 10487.3 & 16.8   & 0.4/0.6 & 4240.3     \\
			Magellan & 7       & 21587.3 & 5.9    & 0.3/0.7 & 8133.2     \\
			Wikidata & 4       & 9489.5  & 15.0   & 0.1/0.9 & 2472.4     \\
			\midrule
			SOTAB-SM & 128     & 254.8   & 13.6   & 0.2/0.8 & 107.2      \\
			\bottomrule
		\end{tabular}
	}
	\caption{
		Average statistics of the benchmark datasets. \#Tasks indicates the number of evaluation pairs in each dataset. \#Rows and \#Cols denote the average number of rows and columns per table. Num/Cat denotes the proportion of numerical and categorical columns, respectively. Cat. Card. denotes the average cardinality of the columns.
	}
	\label{tab:dataset_stats}
\end{table}

\subsection{Datasets}
We evaluate \method{} on the Valentine benchmark~\cite{koutrasValentineEvaluatingMatching2021}, a widely adopted suite for assessing schema matching techniques in the context of dataset discovery. Valentine classifies matching tasks into four distinct relatedness scenarios:

\begin{itemize}
	\item \textbf{Unionable:} Tables that share the same schema and entity type (vertically partitioned).
	\item \textbf{View-Unionable:} Tables that share a subset of attributes but contain unique columns and possess no row overlap.
	\item \textbf{Joinable:} Tables containing complementary information about the same entities, linkable via a key column.
	\item \textbf{Semantically-Joinable:} Joinable tables where the joining columns exhibit semantic heterogeneity or noisy instance representations.
\end{itemize}

The benchmark comprises a total of 540 fabricated dataset pairs and several human-curated real-world pairs. The fabricated datasets are generated from three seed sources: \textit{TPC-DI}~\cite{poess2014tpc} (data integration), \textit{Open Data}~\cite{nargesian2018table} (government census data), and \textit{ChEMBL}~\cite{davies2015chembl} (chemical database). These pairs are systematically perturbed with varying degrees of schema noise (e.g., column renaming, abbreviations) and instance noise (e.g., typos, format variations) to simulate data lake inconsistencies. The real-world component includes curated pairs from \textit{Wikidata}~\cite{vrandevcic2014wikidata} and the \textit{Magellan}~\cite{magellandata} repository, providing ground truth for complex matching scenarios involving natural language variation.~\Cref{tab:dataset_stats} summarizes key statistics of the Valentine benchmark datasets.

While the Valentine benchmark provides a comprehensive set of matching scenarios, it is important to note that the fabrication algorithm is limited -- column headers are mostly similar across source and target tables with only minor perturbations. Thus, to thoroughly evaluate the robustness of \method{}, we additionally evaluate on the SOTAB-SM benchmark, presented as a part of the Web Data Commons Schema Matching Benchmark (WDC-SMB)~\cite{rengWDCSchemaMatching}. The SOTAB-SM benchmark is derived from real-world web tables used in the Schema.org Table Annotation Benchmark (SOTAB)~\cite{koriniSOTABWDCSchemaorg2022}, which was designed to evaluate table annotation systems on Column Type Annotation (CTA) and Column Property Annotation (CPA) tasks. Unlike Valentine, all the tables in SOTAB-SM are sourced from actual websites and annotated according to the Schema.org column properties. The column headers are replaced with simple numbers, making this benchmark a good test for evaluating matching algorithms in real-world noisy settings with opaque column names. We use the \texttt{sotab\_sm\_v500} version, which provides pre-computed train/validation/test splits consisting of 200/200/100 tables. We train~\method{} using both the train and validation tables and evaluate on the test split. In addition to the difficulty of the matching tasks, SOTAB-SM also allows us to understand the generalizability of our method, as the training tables are no longer fabricated from seed datasets but are instead sampled from a large corpus of web tables.

\subsection{Baselines}

\subsubsection{PLM-Based Baselines}
\textbf{Magneto}~\cite{liuMagnetoCombiningSmall2025} is a two-stage framework that utilizes a fine-tuned PLM for initial matching, followed by an LLM-based re-ranking step. To rigorously evaluate the quality of the generated embeddings and ensure a fair comparison of structural encoders, we focus on the first-stage bipartite matching results. We also include the LLM re-ranking results by following the original implementation.
\textbf{ISResMat}~\cite{duSituNeuralRelational2024} is a recent PLM-based schema matching method that fine-tunes a PLM directly on target table pairs. Unlike Magneto, it does not rely on external API calls for re-ranking, making it a direct baseline for embedding-based methods. We utilize the official code repository provided by the authors to generate baseline results.
In addition to the above sophisticated methods, we also implement a simple no-tune \textbf{LM} baseline. This approach uses a frozen PLM to encode columns using a \texttt{verbose} serialization format as described in~\Cref{tab:serialization_desc}. Once we embed each column, we simply compute the softmax-normalized cosine similarity between source and target columns to obtain matching scores, which are then refined by the same bipartite reranking algorithm used in some Magneto variants and~\method{}. While implementation details differ, this approach is essentially the same as \texttt{Magneto-zs-bp} as seen in~\citet{liuMagnetoCombiningSmall2025}.

It is also worth noting that~\citet{koutrasREMAGraphEmbeddingsbased2020} uses a similar graph-based formulation as ours, we were not able to find publicly available code for their implementation, so we do not include it as a baseline. In addition, the node embeddings in REMA are trained from scratch, while our method makes use of pre-trained language models and solely focuses on capturing the interaction between the already defined semantic representations.

In addition, we also consider baselines from pre-LLM era such as COMA++~\cite{aumuellerSchemaOntologyMatching2005}, Jaccard-Distance~\cite{koutrasValentineEvaluatingMatching2021}, Cupid~\cite{madhavan2001generic}, Similarity Flooding~\cite{melnik2002similarity} and Distribution-based Matching~\cite{zhang2011automatic}. We note that these methods generally underperform compared to modern PLM-based approaches, but we include them for completeness.~\Cref{sec:implementation_details} describes the implementation and hardware details for all our experiments.

\section{Results}
\label{sec:results}

\begin{figure}[tp!]
	\centering
	\includegraphics[width=0.9\linewidth]{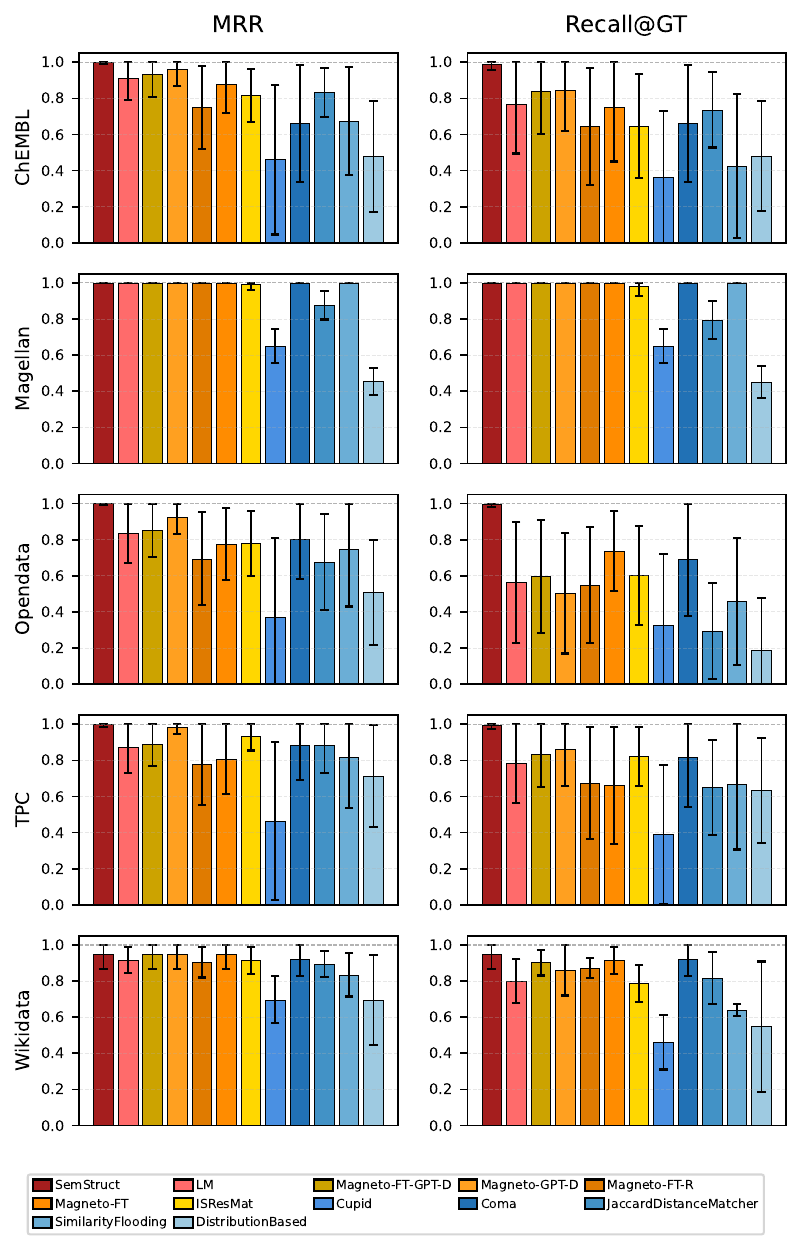}
	\caption{MRR and Recall@GT comparison of different methods on Valentine. The LM baseline already shows strong performance, comparable to previous state-of-the-art methods that require fine-tuning. However, by explicitly modeling structural dependencies,~\method{} achieves significant performance improvements across both metrics, particularly in distinguishing ambiguous columns.}
	\label{fig:main-result-bar}
\end{figure}

\subsection{Overall Results}

\subsubsection{Valentine Benchmark}
\Cref{fig:main-result-bar} presents the results of our proposed \method{} framework compared to state-of-the-art baselines on the Valentine benchmark. A surprising finding is that using frozen, off-the-shelf sentence encoders (specifically E5) can achieve competitive performance against methods requiring fine-tuning (e.g., ISResMat and Magneto-FT). This is surprising given that we are able to outperform these methods without \textit{any} adaptation of the PLM parameters. This also somewhat confirms the findings from~\citet{liuMagnetoCombiningSmall2025}, who noted the strong performance of zero-shot embeddings against the fine-tuned variant in TPC-DI -- however, we find that the LM baseline also outperforms the fine-tuned Magneto in ChEMBL as well. This discrepancy may be due to the different PLM backbone used -- Magneto-FT uses MPNet (which was reported as the best performer), while we identify E5 as the best overall performer for our LM baseline and~\method{}. The only difference between our LM baseline and Magneto-zs-bp should be in the serialization of the columns. We conjecture that this improvement in result is due to the effectiveness of the \texttt{verbose} serialization format used, which again highlights the sensitive nature of serialization strategies for tabular data.

However, semantic priors alone are insufficient for complex schema matching. By incorporating structural information through our GNN layers, \method{} achieves statistically significant improvements in both MRR and Recall@GT across all datasets. We also note that the Magellan dataset is particularly \textit{easy}, with both our LM baseline and~\method{} achieving perfect performance. Thus, we remove this dataset from the ablation analysis.

\subsubsection{SOTAB-SM Benchmark}

\begin{figure}[t!]
	\centering
	\includegraphics[width=0.9\linewidth]{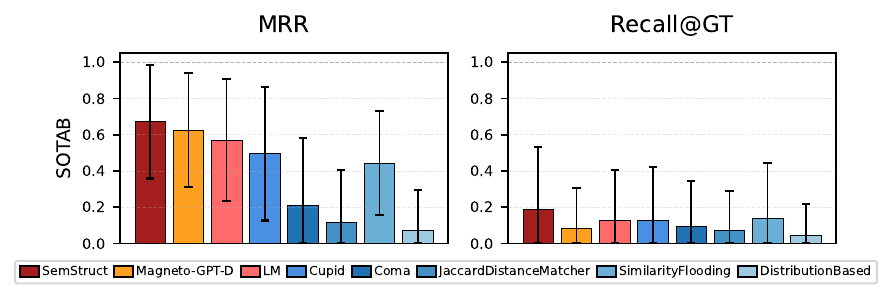}
	\caption{
		Comparison of different methods on the SOTAB-SM benchmark.~\method{} outperforms all baselines in both MRR and Recall@GT, demonstrating its robustness in real-world noisy settings with opaque column headers.
	}
	\label{fig:sotab-result-bar}
\end{figure}

As noted in~\Cref{sec:experiments}, the SOTAB-SM benchmark comprises more challenging real-world tables that are missing column headers and contain more columns with higher semantic content in the cell values (e.g. columns with plain text values such as ``review''). This makes SOTAB-SM an excellent testbed for evaluating the robustness of~\method{}, as well as the utility of the GNN encoder in capturing semantic relationships from the values alone. We exclude Magneto-FT from this evaluation as it requires fine-tuning on LLM-generated data, and ISResMat due to issues in the implementation causing errors on majority of the evaluations due to the token intensive serialization method and the verbose nature of SOTAB columns. Again, we find that~\method{} outperforms all baselines in both MRR and Recall@GT, confirming the effectiveness of our approach in real-world noisy settings.

\subsection{Domain Adaptation vs. Structural Reasoning}

\begin{figure}[tp!]
	\centering
	\includegraphics[width=0.9\linewidth]{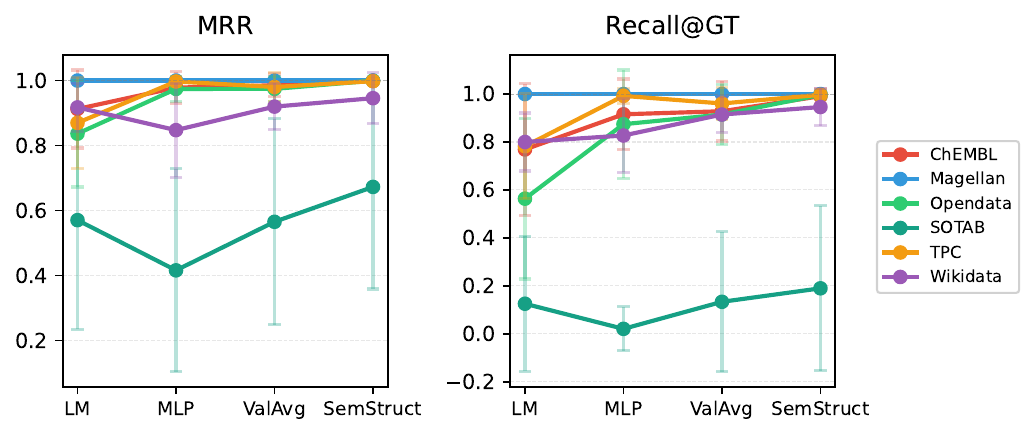}
	\caption{
		Performance evolution as we progressively add model components. \textbf{MLP} (Domain Adaptation) improves over raw embeddings by aligning the vector space. \textbf{ValAvg} (Naive Context) improves further by incorporating value signals. \method{} (Structural Reasoning) yields the highest performance, confirming that explicit message passing is superior to simple averaging for capturing row-level dependencies.
	}
	\label{fig:struct-evolve-line}
\end{figure}

To validate our hypothesis that \textit{structural reasoning} (via GNNs) is superior to simple \textit{context aggregation}, we conduct a component-wise ablation study (\Cref{fig:struct-evolve-line}). We define three progressive configurations:
\begin{enumerate}
	\item \textbf{MLP (Domain Adaptation):} A trained 2-layer MLP on top of frozen column node embeddings. This isolates the effect of learning a task-specific projection head.
	\item \textbf{ValAvg (Naive Context):} Instead of using only the column node, we average the embeddings of the unique values in that column and pass this centroid to the MLP. This represents a primitive, non-relational way of injecting data context.
	\item \textbf{\method{} (Structural Reasoning):} The complete framework utilizing the heterogeneous graph and GNN message passing.
\end{enumerate}

For most Valentine datasets, we observe a clear hierarchy of performance. While \textbf{MLP} provides a necessary alignment of the embedding space, and \textbf{ValAvg} offers a lift by introducing instance data, both are outperformed by \method{}. It is worth noting that this trend is observed in three out of four datasets in the Valentine benchmark, while MLP-tuning on SOTAB and Wikidata actually shows worse performance than the frozen PLM. We hypothesize that this is due to the smaller vocabulary size of the synthetically generated Valentine datasets, which allows the MLP layer to learn effective projections for those constrained domains. In contrast, SOTAB and Wikidata's cross-topic nature and long text values make it more difficult to learn an effective adaptation without overfitting.

\begin{figure}[tp!]
	\centering
	\includegraphics[width=\linewidth]{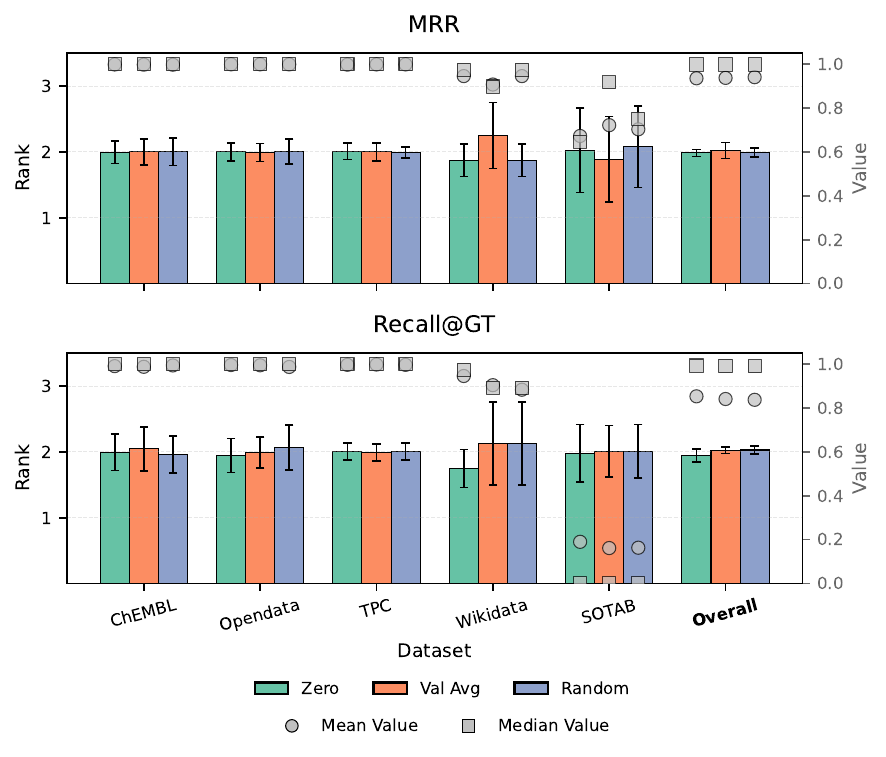}
	\caption{Comparison of row node initialization strategies. Lower rank means better performance (left axis). Mean and median scores are shown in scatter points (right axis). Zero-initialization outperforms both value-averaging and random initialization. This supports the hypothesis that row nodes act as topological bridges rather than semantic entities.}
	\label{fig:row-init-ablation}
\end{figure}

The \textit{usefulness} of structural reasoning becomes apparent when comparing~\method{} against \textbf{ValAvg}. \textbf{ValAvg} adds all unique values' embeddings into a single mean embedding, effectively creating a ``semantic centroid'' for that column. If a column contains distinct, heterogeneous values (e.g., integers and strings), averaging them dilutes the signal, resulting in a noisy representation (``semantic mud''). In contrast, \method{} maintains distinct nodes for values and uses the row nodes to propagate signals. This allows the model to learn that a specific value (e.g., ``London'') implies a specific column type (e.g., ``City'') via its row-level co-occurrence with other entities, without that signal being washed out by the average of the entire column. For all of the datasets, we observe an upwards trend in both metrics when moving from \textbf{ValAvg} to \method{}, confirming that explicit structural reasoning adds something beyond naive context aggregation.

\begin{figure}[tp!]
	\centering
	\includegraphics[width=\linewidth]{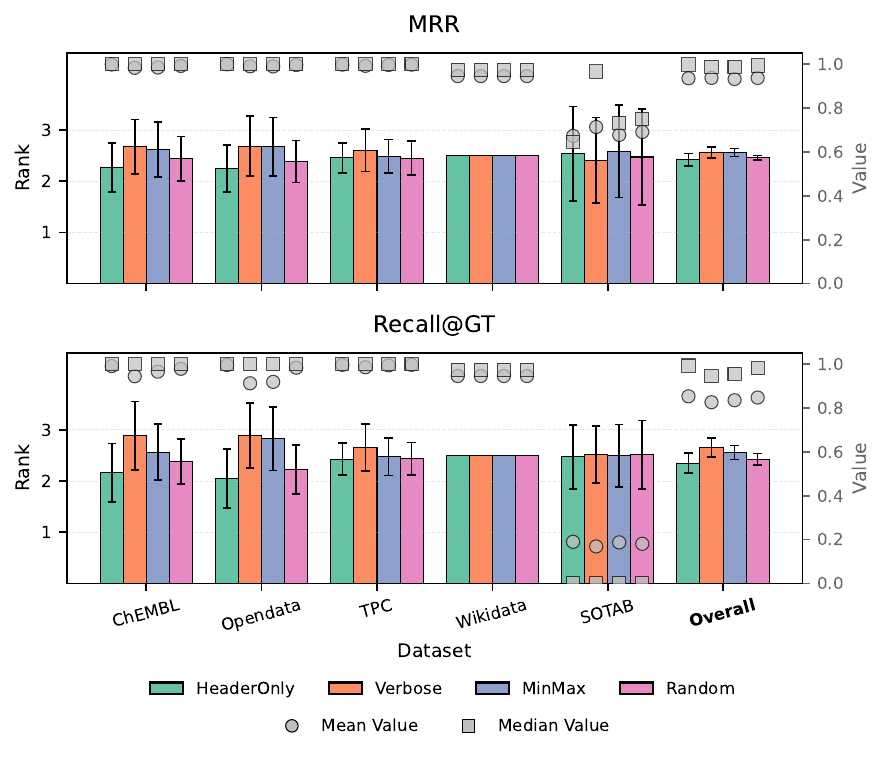}
	\caption{
		Comparison of different serialization strategies. Lower rank means better performance (left axis). Mean and median scores are shown in scatter points (right axis). \texttt{header} serialization offers (near) best performance, even in SOTAB benchmark where the column names are opaque, suggesting that the GNN encoder is able to effectively propagate structural context to the column nodes.
	}
	\label{fig:serialization-ablation}
\end{figure}

\subsection{Row Node Initialization Strategies}
\label{sec:reseults-row-init}

In our heterogeneous graph, Column and Value nodes possess inherent semantic information derived from the PLM. However, Row nodes represent abstract tuples; they have no textual representation. This raises the question: how should we initialize row nodes to best facilitate message passing?

We evaluated three strategies: (1) \textbf{Zero-Initialization}, (2) \textbf{Value Averaging} (initializing the row as the mean of its constituent values), and (3) \textbf{Random Initialization}. \textbf{Zero-Initialization} treats the row node as just an \textit{information conduit}, while \textbf{Value Averaging} attempts to imbue the row with semantic content and also give each row a notion of \textit{identity}. \textbf{Random Initialization} serves as a control to test whether any non-zero signal (\textit{identity} alone) is beneficial.

As shown in \Cref{fig:row-init-ablation}, we find that \textbf{Zero-Initialization} shows the best overall performance, especially notable in Wikidata. However, on the more realistic SOTAB, we find that \textbf{Value Averaging} and \textbf{Random Initialization} perform comparably in Recall@GT, with stronger MRR when compared to \textbf{Zero-Initialization}. This suggests that in real-world noisy tables, providing some form of identity or semantic signal to the row nodes can be beneficial.

\subsection{Column Node Serialization}

Next, we investigate how different serialization strategies impact performance. As discussed in~\Cref{tab:serialization_desc}, we test four variants:
\texttt{header}, \texttt{verbose}, \texttt{minmax}, and \texttt{random}.~\Cref{fig:serialization-ablation} shows the average rank of each serialization method across all datasets. Surprisingly, we find that \texttt{header} serialization is a very strong contender, resulting in the lowest rank (best performance) across almost all datasets and metrics. This is particularly notable in the SOTAB benchmark, where column headers are replaced with just numbers (e.g., ``0'', ``1''). The strong performance of \texttt{header} serialization in this scenario suggests that the GNN is effectively propagating the context of the values to the column nodes, allowing them to disambiguate based on structural co-occurrences rather than relying solely on the header text or column-wise value summaries.

\subsection{Impact of Semantic Encoders}

\begin{figure}
	\centering
	\includegraphics[width=\linewidth]{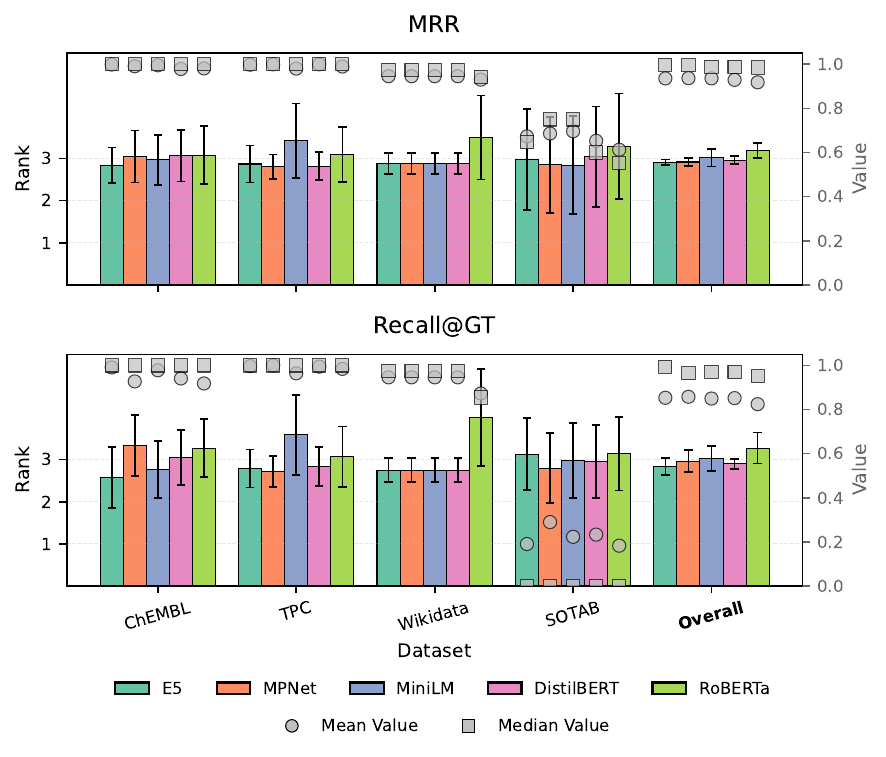}
	\caption{
		Comparison of different PLM backbones. Lower rank means better performance (left axis). Mean and median scores are shown in scatter points (right axis). E5-small generally shows the best performance, but MPNet shows the best performance on SOTAB-SM.
	}
	\label{fig:lm-backbone-ablation}
\end{figure}

Finally, we evaluate the impact of different PLM backbones on performance. We compare five popular lightweight sentence encoders: E5-small~\cite{wangTextEmbeddingsWeaklySupervised2024}, MPNet~\cite{songMPNetMaskedPermuted2020}, MiniLM~\cite{wangMiniLMDeepSelfAttention2020}, RoBERTa-base~\cite{liuRoBERTaRobustlyOptimized2019} and distilBERT~\cite{sanhDistilBERTDistilledVersion2020}. As shown in \Cref{fig:lm-backbone-ablation}, we find that E5-small generally offers the best performance across most datasets, likely due to its training on a large corpus of diverse text pairs. However, on the SOTAB-SM benchmark, MPNet slightly outperforms E5-small. This suggests that MPNet's masked and permuted language modeling objectives may better capture the nuanced semantics present in real-world web tables. Overall, these results indicate that while the choice of PLM backbone does impact performance, the structural reasoning provided by the GNN encoder remains the dominant factor in achieving high schema matching accuracy.

\subsection{Case Study: Cross-Domain Schema Matching}

\begin{table}
	\centering
	{\footnotesize
		\begin{tabular}{lrll}
			\toprule
			Method                     & Score & Example Values                                                 &        \\
			\midrule
			Source                     & 1.000 & ``Shake Things Up Cocktail Basket'', ``Bistro Chic'',~\ldots   &        \\
			\midrule
			\multirow{5}{*}{LM}        & 0.790 & ``5'', ``5'', ``5'', ``5'', ``5'',~\ldots                      & \xmark \\
			                           & 0.739 & ``4217'', ``4217'', ``4217'', ``4217'', ``4217'',~\ldots       & \xmark \\
			                           & 0.733 & ``-27.3819'', ``-34.4283'', ``-26.6064'', ``-32.9766'',~\ldots & \xmark \\
			                           & 0.733 & ``Thursday Friday Monday Tuesday Wednesday'',~\ldots           & \xmark \\
			                           & 0.729 & ``1300 032 861'', ``1300 032 861'', ``1300 032 861'',~\ldots   & \xmark \\
			\midrule
			\multirow{5}{*}{\method{}} & 0.614 & ``Bricklayers Near Me'', ``Bricklayers Near Me'',~\ldots       & \cmark \\
			                           & 0.303 & ``QLD'', ``QLD'', ``QLD'', ``QLD'', ``QLD''                    & \xmark \\
			                           & 0.252 & ``5'', ``5'', ``5'', ``5'', ``5'',~\ldots                      & \xmark \\
			                           & 0.243 & ``5'', ``5'', ``5'', ``5'', ``5'',~\ldots                      & \xmark \\
			                           & 0.203 & ``1300 032 861'', ``1300 032 861'', ``1300 032 861'',~\ldots   & \xmark \\
			\bottomrule
		\end{tabular}
	}
	\caption{
		Example of top-5 matched columns from a SOTAB-SM task using LM and~\method{}. The LM is not able to tell apart columns with ambiguous values, assigning high similarity scores to multiple incorrect candidates, while the~\method{} is able to correctly identify the target column while assigning lower scores to distractors.
	}
	\label{tab:case-study-sotab}
\end{table}

To better understand how~\method{} successfully leverages structural reasoning where the traditional LM baseline fails, we examine examples from the SOTAB-SM benchmark, which provides the most challenging real-world tasks by removing column headers and having many semantically ambiguous columns (repeated numbers, code words, etc.).~\Cref{tab:case-study-sotab} shows an example of a matching task between Schema.org's \texttt{Product} and \texttt{LocalBusiness} types, where the task is to match the \texttt{Product.name} column (Source) to \texttt{LocalBusiness.name} (Target). As shown in the table, the LM baseline struggles to disambiguate between multiple columns with semantically ambiguous content. This phenomenon is also known as the \textit{embedding anisotropy} problem~\cite{ethayarajhHowContextualAre2019}, where the embedding space used by the language model is clustered around certain high-density regions, causing high similarity even for inputs that are not truly related. In our example, the LM also assigns high similarity score to multiple columns that are seemingly not related, while~\method{} only assigns a high score to the correct target column and lower scores to all other candidates. This is especially noteworthy considering that~\method{} is \textit{adding} the structural embedding on top of the same frozen LM embeddings, indicating that the GNN is effectively refining the representations to better capture the relational context.

\section{Conclusion}

In this paper, we introduced~\method{}, a novel framework for schema matching that combines the semantic power of frozen pre-trained language models (PLMs) with the structural reasoning capabilities of Graph Neural Networks (GNNs). By explicitly modeling row-level value co-occurrences through a heterogeneous graph structure,~\method{} effectively captures the instance-level interactions between attributes that traditional serialized approaches overlook. Our experiments on the Valentine benchmark and the SOTAB-SM dataset demonstrate that~\method{} achieves state-of-the-art performance without using any proprietary LLMs or fine-tuning large language models, outperforming existing methods. Our ablation studies show that the graph formulation is a key contributor to the performance gains. In addition, we show that row nodes primarily serve as structural conduits, highlighting the importance of topological information in schema matching tasks. Overall,~\method{} presents a scalable solution for schema matching in heterogeneous data environments, paving the way for future research in integrating semantic and structural representations.

Future work can extend our work by considering more complex scenarios, such as incorporating external information not available in the table, which is crucial for many industry use cases. Additionally, in cases where the tables contain sensitive information, it is also worth exploring how to adapt our proposed approach to work with privacy-preserving techniques.

\section{Acknowledgments}
This work was supported by IBM through the IBM-Rensselaer Future of Computing Research Collaboration.

\bibliographystyle{ACM-Reference-Format}
\bibliography{references}

\appendix
\section{Appendix}

\subsection{Experimental Setup}
\label{sec:implementation_details}

\subsubsection{Hardware \& Libraries}
We utilize PyTorch Geometric~\cite{feyPyG20Scalable2025} for graph-based operations and the Hugging Face Transformers library~\cite{wolfHuggingFacesTransformersStateoftheart2020} for language model integrations. All experiments were conducted on a system equipped with a single NVIDIA H100 GPU. We use the E5~\cite{wangTextEmbeddingsWeaklySupervised2024} as the default embedding model unless otherwise specified.

\subsubsection{Training \& Hyperparameters}
\method{} is trained using the Adam optimizer. For the structural encoder, we employ a 3-layer GraphSAGE~\cite{hamiltonInductiveRepresentationLearning2017} backbone with a hidden dimension of 384. For the semantic initialization, we utilize the frozen \texttt{e5-small}~\cite{wangTextEmbeddingsWeaklySupervised2024} checkpoint. The model is trained for 10 epochs with a batch size of 16 tables and learning rate of $5\text{e-}4$ for Valentine, and 30 epochs and a learning rate of $1\text{e-}4$ for SOTAB-SM. We find that the smaller size of the training data for SOTAB-SM benefits from longer training. We also use the validation set for early stopping, stopping if there are no improvements in the validation loss for 5 consecutive epochs.

We use $k=8$, the number of bins used in our discretization steps, throughout all our experiments. This value was chosen after a rough hyperparameter sweep on the validation data, where we noted that larger values of $k$ yield diminishing and sometimes even worse returns.

\subsection{Detailed Experimental Results}

Here we provide the detailed experimental results for all the ablation studies considered in the main paper. Each of~\Cref{tab:row_init_mrr,tab:row_init_recall_gt,tab:serialization_mrr,tab:serialization_recall_gt,tab:lm_backbone_mrr,tab:lm_backbone_recall_gt} shows the median value and interquartile range of the individual metrics for each dataset. Best performance and ranks are highlighted in \textbf{bold}, followed by the second best in \underline{underline}.

\setlength{\tabcolsep}{2pt}

\begin{table}[h]
	\centering
	\caption{Comparing different row initialization schemes. (MRR)}
	\label{tab:row_init_mrr}
	{\small
		\begin{tabular}{lccc}
			\toprule
			Dataset  & Zero                                       & Val Avg                                          & Random                                           \\
			\midrule
			ChEMBL   & 1.000{\tiny(0.000)}/\textbf{2.00}          & 1.000{\tiny(0.000)}/\underline{2.00}             & 1.000{\tiny(0.000)}/2.00                         \\
			Opendata & 1.000{\tiny(0.000)}/\underline{2.00}       & 1.000{\tiny(0.000)}/\textbf{1.99}                & 1.000{\tiny(0.000)}/2.01                         \\
			TPC      & 1.000{\tiny(0.000)}/2.01                   & 1.000{\tiny(0.000)}/\underline{2.00}             & 1.000{\tiny(0.000)}/\textbf{1.99}                \\
			Wikidata & \textbf{0.975}{\tiny(0.079)}/\textbf{1.88} & \underline{0.898}{\tiny(0.120)}/\underline{2.25} & \textbf{0.975}{\tiny(0.079)}/\textbf{1.88}       \\
			SOTAB    & 0.646{\tiny(0.500)}/\underline{2.03}       & \textbf{0.918}{\tiny(0.500)}/\textbf{1.89}       & \underline{0.750}{\tiny(0.500)}/2.08             \\
			\midrule
			Overall  & 0.998{\tiny(0.040)}/\textbf{1.98}          & \textbf{0.999}{\tiny(0.069)}/2.02                & \underline{0.999}{\tiny(0.040)}/\underline{1.99} \\
			\bottomrule
		\end{tabular}
	}
\end{table}

\begin{table}[h]
	\centering
	\caption{Comparing different row initialization schemes. (Recall@GT)}
	\label{tab:row_init_recall_gt}
	{\small
		\begin{tabular}{lccc}
			\toprule
			Dataset  & Zero                                       & Val Avg                                          & Random                                           \\
			\midrule
			ChEMBL   & 1.000{\tiny(0.000)}/\underline{1.99}       & 1.000{\tiny(0.000)}/2.04                         & 1.000{\tiny(0.000)}/\textbf{1.96}                \\
			Opendata & 1.000{\tiny(0.000)}/\textbf{1.94}          & 1.000{\tiny(0.000)}/\underline{1.99}             & 1.000{\tiny(0.000)}/2.07                         \\
			TPC      & 1.000{\tiny(0.000)}/\underline{2.01}       & 1.000{\tiny(0.000)}/\textbf{1.99}                & 1.000{\tiny(0.000)}/\underline{2.01}             \\
			Wikidata & \textbf{0.975}{\tiny(0.079)}/\textbf{1.75} & \underline{0.892}{\tiny(0.129)}/\underline{2.12} & \underline{0.892}{\tiny(0.150)}/\underline{2.12} \\
			SOTAB    & 0.000{\tiny(0.333)}/\textbf{1.98}          & 0.000{\tiny(0.000)}/\underline{2.01}             & 0.000{\tiny(0.187)}/\underline{2.01}             \\
			\midrule
			Overall  & \textbf{0.993}{\tiny(0.039)}/\textbf{1.95} & \underline{0.991}{\tiny(0.072)}/\underline{2.03} & 0.991{\tiny(0.087)}/2.03                         \\
			\bottomrule
		\end{tabular}
	}
\end{table}

\begin{table}[h]
	\centering
	\caption{Comparing different column node serialization schemes. (MRR)}
	\label{tab:serialization_mrr}
	{\footnotesize
		\begin{tabular}{lcccc}
			\toprule
			Dataset  & HeaderOnly                                 & Verbose                                    & MinMax                   & Random                                           \\
			\midrule
			ChEMBL   & 1.000{\tiny(0.000)}/\textbf{2.27}          & 1.000{\tiny(0.000)}/2.67                   & 1.000{\tiny(0.000)}/2.62 & 1.000{\tiny(0.000)}/\underline{2.44}             \\
			Opendata & 1.000{\tiny(0.000)}/\textbf{2.25}          & 1.000{\tiny(0.000)}/2.69                   & 1.000{\tiny(0.000)}/2.67 & 1.000{\tiny(0.000)}/\underline{2.39}             \\
			TPC      & 1.000{\tiny(0.000)}/\underline{2.46}       & 1.000{\tiny(0.000)}/2.61                   & 1.000{\tiny(0.000)}/2.49 & 1.000{\tiny(0.000)}/\textbf{2.45}                \\
			Wikidata & 0.975{\tiny(0.079)}/2.50                   & 0.975{\tiny(0.079)}/2.50                   & 0.975{\tiny(0.079)}/2.50 & 0.975{\tiny(0.079)}/2.50                         \\
			SOTAB    & 0.646{\tiny(0.500)}/2.54                   & \textbf{0.967}{\tiny(0.500)}/\textbf{2.41} & 0.732{\tiny(0.548)}/2.58 & \underline{0.750}{\tiny(0.500)}/\underline{2.47} \\
			\midrule
			Overall  & \textbf{0.998}{\tiny(0.040)}/\textbf{2.42} & 0.987{\tiny(0.037)}/2.56                   & 0.987{\tiny(0.039)}/2.56 & \underline{0.994}{\tiny(0.040)}/\underline{2.46} \\
			\bottomrule
		\end{tabular}
	}
\end{table}

\begin{table}[h]
	\centering
	\caption{Comparing different column node serialization schemes. (Recall@GT)}
	\label{tab:serialization_recall_gt}
	{\footnotesize
		\begin{tabular}{lcccc}
			\toprule
			Dataset  & HeaderOnly                                 & Verbose                  & MinMax                               & Random                                           \\
			\midrule
			ChEMBL   & 1.000{\tiny(0.000)}/\textbf{2.17}          & 1.000{\tiny(0.091)}/2.89 & 1.000{\tiny(0.062)}/2.57             & 1.000{\tiny(0.000)}/\underline{2.38}             \\
			Opendata & 1.000{\tiny(0.000)}/\textbf{2.05}          & 1.000{\tiny(0.067)}/2.89 & 1.000{\tiny(0.060)}/2.83             & 1.000{\tiny(0.000)}/\underline{2.23}             \\
			TPC      & 1.000{\tiny(0.000)}/\textbf{2.43}          & 1.000{\tiny(0.000)}/2.66 & 1.000{\tiny(0.000)}/2.48             & 1.000{\tiny(0.000)}/\underline{2.44}             \\
			Wikidata & 0.975{\tiny(0.079)}/2.50                   & 0.975{\tiny(0.079)}/2.50 & 0.975{\tiny(0.079)}/2.50             & 0.975{\tiny(0.079)}/2.50                         \\
			SOTAB    & 0.000{\tiny(0.333)}/\textbf{2.47}          & 0.000{\tiny(0.213)}/2.52 & 0.000{\tiny(0.259)}/\underline{2.49} & 0.000{\tiny(0.042)}/2.52                         \\
			\midrule
			Overall  & \textbf{0.993}{\tiny(0.039)}/\textbf{2.35} & 0.945{\tiny(0.055)}/2.66 & 0.956{\tiny(0.061)}/2.56             & \underline{0.981}{\tiny(0.039)}/\underline{2.43} \\
			\bottomrule
		\end{tabular}
	}
\end{table}

\begin{table}[h]
	\centering
	\caption{Comparing different LLM backbones. (MRR)}
	\label{tab:lm_backbone_mrr}
	{\tiny
		\begin{tabular}{lccccc}
			\toprule
			Dataset  & E5                                         & MPNet                                            & MiniLM                                     & DistilBERT                                 & RoBERTa                                          \\
			\midrule
			ChEMBL   & 1.000{\tiny(0.000)}/\textbf{2.84}          & 1.000{\tiny(0.000)}/3.05                         & 1.000{\tiny(0.000)}/\underline{2.97}       & 1.000{\tiny(0.000)}/3.07                   & 1.000{\tiny(0.000)}/3.08                         \\
			Opendata & 1.000{\tiny(0.000)}/\textbf{2.91}          & 1.000{\tiny(0.000)}/\underline{2.93}             & 1.000{\tiny(0.000)}/2.99                   & 1.000{\tiny(0.000)}/2.98                   & 1.000{\tiny(0.000)}/3.19                         \\
			TPC      & 1.000{\tiny(0.000)}/2.87                   & 1.000{\tiny(0.000)}/\textbf{2.80}                & 1.000{\tiny(0.000)}/3.42                   & 1.000{\tiny(0.000)}/\underline{2.82}       & 1.000{\tiny(0.000)}/3.09                         \\
			Wikidata & \textbf{0.975}{\tiny(0.079)}/\textbf{2.88} & \textbf{0.975}{\tiny(0.079)}/\textbf{2.88}       & \textbf{0.975}{\tiny(0.079)}/\textbf{2.88} & \textbf{0.975}{\tiny(0.079)}/\textbf{2.88} & \underline{0.944}{\tiny(0.051)}/\underline{3.50} \\
			SOTAB    & \underline{0.646}{\tiny(0.500)}/2.98       & \textbf{0.750}{\tiny(0.604)}/\underline{2.85}    & \textbf{0.750}{\tiny(0.500)}/\textbf{2.84} & 0.600{\tiny(0.667)}/3.04                   & 0.552{\tiny(0.708)}/3.29                         \\
			\midrule
			Overall  & \textbf{0.998}{\tiny(0.040)}/\textbf{2.91} & \underline{0.995}{\tiny(0.043)}/\underline{2.92} & 0.988{\tiny(0.043)}/3.02                   & 0.989{\tiny(0.045)}/2.96                   & 0.985{\tiny(0.052)}/3.19                         \\
			\bottomrule
		\end{tabular}
	}
\end{table}

\begin{table}[h]
	\centering
	\caption{Comparing different LLM backbones. (Recall@GT)}
	\label{tab:lm_backbone_recall_gt}
	{\tiny
		\begin{tabular}{lccccc}
			\toprule
			Dataset  & E5                                         & MPNet                                      & MiniLM                                     & DistilBERT                                 & RoBERTa                                          \\
			\midrule
			ChEMBL   & 1.000{\tiny(0.000)}/\textbf{2.57}          & 1.000{\tiny(0.091)}/3.33                   & 1.000{\tiny(0.000)}/\underline{2.77}       & 1.000{\tiny(0.062)}/3.05                   & 1.000{\tiny(0.062)}/3.27                         \\
			Opendata & 1.000{\tiny(0.000)}/\textbf{2.77}          & 1.000{\tiny(0.000)}/3.20                   & 1.000{\tiny(0.000)}/3.07                   & 1.000{\tiny(0.000)}/\underline{2.79}       & 1.000{\tiny(0.000)}/3.17                         \\
			TPC      & 1.000{\tiny(0.000)}/\underline{2.79}       & 1.000{\tiny(0.000)}/\textbf{2.72}          & 1.000{\tiny(0.067)}/3.59                   & 1.000{\tiny(0.000)}/2.84                   & 1.000{\tiny(0.000)}/3.07                         \\
			Wikidata & \textbf{0.975}{\tiny(0.079)}/\textbf{2.75} & \textbf{0.975}{\tiny(0.079)}/\textbf{2.75} & \textbf{0.975}{\tiny(0.079)}/\textbf{2.75} & \textbf{0.975}{\tiny(0.079)}/\textbf{2.75} & \underline{0.854}{\tiny(0.060)}/\underline{4.00} \\
			SOTAB    & 0.000{\tiny(0.333)}/3.13                   & 0.000{\tiny(0.500)}/\textbf{2.80}          & 0.000{\tiny(0.500)}/2.98                   & 0.000{\tiny(0.500)}/\underline{2.95}       & 0.000{\tiny(0.175)}/3.14                         \\
			\midrule
			Overall  & \textbf{0.993}{\tiny(0.039)}/\textbf{2.84} & 0.965{\tiny(0.064)}/2.97                   & \underline{0.971}{\tiny(0.035)}/3.03       & 0.970{\tiny(0.054)}/\underline{2.90}       & 0.951{\tiny(0.102)}/3.28                         \\
			\bottomrule
		\end{tabular}
	}
\end{table}

\begin{table}[h]
	\centering
	\caption{MRR by Match Type (Median / IQR)}
	\label{tab:mrr_by_match_type}
	{\footnotesize
		\begin{tabular}{lcccc}
			\toprule
			Matcher            & Joinable                        & Sem.-Joinable                   & Unionable                       & View-Unionable                  \\
			\midrule
			SemStruct          & \textbf{1.000}{\tiny(0.000)}    & \textbf{1.000}{\tiny(0.000)}    & \textbf{1.000}{\tiny(0.000)}    & \textbf{1.000}{\tiny(0.000)}    \\
			Coma               & \textbf{1.000}{\tiny(0.200)}    & 0.867{\tiny(0.409)}             & 0.804{\tiny(0.366)}             & 0.826{\tiny(0.409)}             \\
			Cupid              & 0.250{\tiny(0.962)}             & 0.167{\tiny(1.000)}             & 0.174{\tiny(0.887)}             & 0.250{\tiny(0.971)}             \\
			DistributionBased  & 0.903{\tiny(0.262)}             & 0.428{\tiny(0.386)}             & 0.589{\tiny(0.122)}             & 0.348{\tiny(0.474)}             \\
			ISResMat           & \underline{0.967}{\tiny(0.146)} & 0.791{\tiny(0.219)}             & 0.838{\tiny(0.117)}             & 0.842{\tiny(0.316)}             \\
			Jaccard            & \textbf{1.000}{\tiny(0.160)}    & 0.812{\tiny(0.340)}             & 0.768{\tiny(0.116)}             & 0.755{\tiny(0.269)}             \\
			LM                 & \textbf{1.000}{\tiny(0.069)}    & 0.794{\tiny(0.289)}             & \underline{0.925}{\tiny(0.187)} & 0.891{\tiny(0.269)}             \\
			Magneto-FT         & 0.915{\tiny(0.288)}             & 0.889{\tiny(0.303)}             & 0.843{\tiny(0.390)}             & 0.840{\tiny(0.378)}             \\
			SimilarityFlooding & 0.960{\tiny(0.500)}             & \underline{0.943}{\tiny(0.483)} & 0.864{\tiny(0.522)}             & \underline{0.955}{\tiny(0.500)} \\
			\bottomrule
		\end{tabular}
	}
\end{table}

\begin{table}[h]
	\centering
	\caption{Recall@GT by Match Type (Median / IQR)}
	\label{tab:recall_by_match_type}
	{\footnotesize
		\begin{tabular}{lcccc}
			\toprule
			Matcher            & Joinable                        & Sem.-Joinable                   & Unionable                       & View-Unionable                  \\
			\midrule
			SemStruct          & \textbf{1.000}{\tiny(0.000)}    & \textbf{1.000}{\tiny(0.000)}    & \textbf{1.000}{\tiny(0.000)}    & \textbf{1.000}{\tiny(0.000)}    \\
			Coma               & \underline{0.986}{\tiny(0.292)} & 0.721{\tiny(0.520)}             & 0.804{\tiny(0.366)}             & \underline{0.746}{\tiny(0.544)} \\
			Cupid              & 0.167{\tiny(0.805)}             & 0.167{\tiny(0.830)}             & 0.189{\tiny(0.636)}             & 0.200{\tiny(0.805)}             \\
			DistributionBased  & 0.815{\tiny(0.610)}             & 0.333{\tiny(0.451)}             & 0.565{\tiny(0.562)}             & 0.167{\tiny(0.475)}             \\
			ISResMat           & 0.833{\tiny(0.200)}             & 0.553{\tiny(0.282)}             & 0.784{\tiny(0.077)}             & 0.667{\tiny(0.357)}             \\
			Jaccard            & 0.833{\tiny(0.444)}             & 0.500{\tiny(0.467)}             & 0.636{\tiny(0.447)}             & 0.438{\tiny(0.448)}             \\
			LM                 & 0.867{\tiny(0.273)}             & 0.625{\tiny(0.436)}             & \underline{0.909}{\tiny(0.227)} & 0.724{\tiny(0.367)}             \\
			Magneto-FT         & 0.738{\tiny(0.400)}             & \underline{0.750}{\tiny(0.400)} & 0.803{\tiny(0.409)}             & 0.662{\tiny(0.560)}             \\
			SimilarityFlooding & 0.456{\tiny(0.809)}             & 0.455{\tiny(0.818)}             & 0.602{\tiny(0.735)}             & 0.400{\tiny(0.878)}             \\
			\bottomrule
		\end{tabular}
	}
\end{table}

\Cref{tab:mrr_by_match_type,tab:recall_by_match_type} present the MRR and Recall@GT results broken down by match type for all baselines and~\method{} on the Valentine benchmark using median and IQR. Two things are notable --~\method{} achieves near perfect scores across all match types, demonstrating its robustness even in challenging semantically-joinable scenarios. Additionally, traditional methods like Coma and Similarity Flooding still perform reasonably well on simpler joinable and unionable tasks, indicating that heuristic approaches retain some efficacy in less complex settings.

\subsection{Impact of Aggregation Layers}

The GNN layer used in~\method{} can influence how information is propagated through the Row-Value-Column graph we construct. In our experiments, we compare three architectures: GraphSAGE~\cite{hamiltonInductiveRepresentationLearning2017}, Graph Attention Networks (GAT)~\cite{velickovicGraphAttentionNetworks2018}, and Heterogeneous Graph Transformers (HGT)~\cite{huHeterogeneousGraphTransformer2020}. Each architecture embodies different inductive biases: GraphSAGE uses fixed-size neighborhood sampling and mean aggregation, GAT employs attention mechanisms to weigh neighbor contributions, and HGT extends attention to heterogeneous graphs with type-specific parameters.

We find that GraphSAGE is not only the most performant but also the most scalable by far. In our experiments with larger tables from the Valentine benchmark, attention-based mechanisms (GAT, HGT) frequently encountered Out-Of-Memory (OOM) errors on a 96 GB VRAM GPU due to the quadratic complexity of attention scores with respect to node degree (number of rows/values).
On the other hand, GraphSAGE, which relies on fixed-size neighborhood sampling and mean aggregation, scales linearly. The superior performance of GraphSAGE suggests that for tabular data -- which often forms a dense bipartite-like structure -- complex attention mechanisms are unnecessary. Simple, isotropic neighborhood aggregation is sufficient to capture the co-occurrence signals required for schema matching. We were not able to complete the comparison between the three because of the memory issues, but even the memory usage alone is enough to demonstrate simple neighbor-based aggregations like GraphSAGE is sufficient.

\subsection{Impact of Value Node Count}

\begin{figure}
	\centering
	\includegraphics[width=0.9\linewidth]{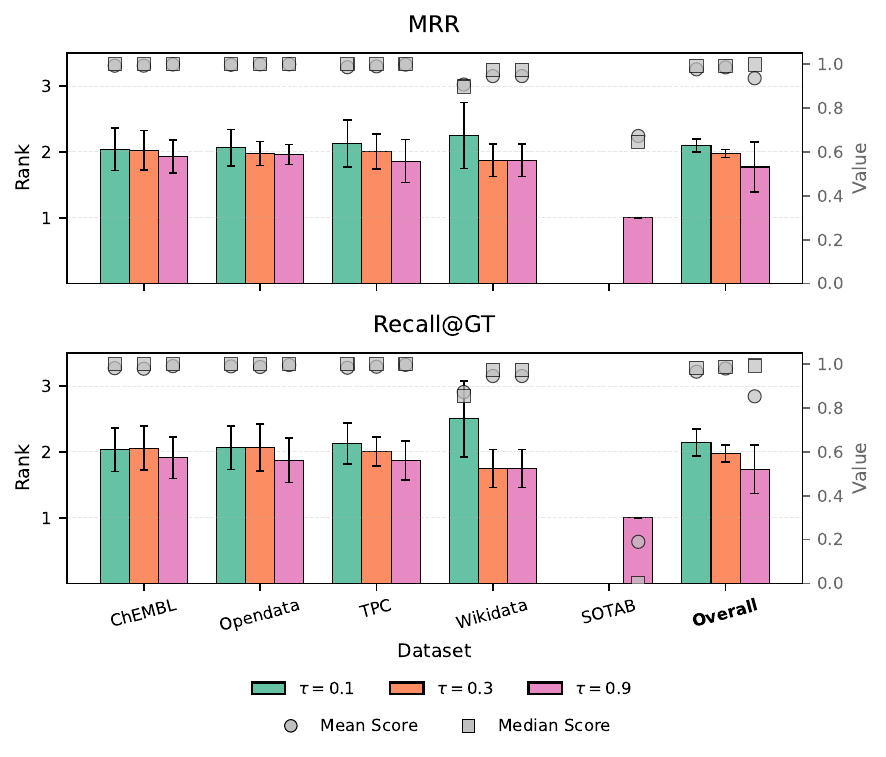}
	\caption{
		Comparison of different node-merging strategies.
	}
	\label{fig:node_merge}
\end{figure}

To understand the impact of the value node construction, we conduct an ablation study where we vary the threshold $\tau$ for merging value nodes based on their co-occurrence patterns. Specifically, we experiment with $\tau \in \{0.0, 0.25, 0.5, 0.75\}$, where $\tau=0.0$ corresponds to no merging (each unique value is its own node) and $\tau=1.0$ corresponds to merging all values into a single node. We find that even using a single node for all values yields competitive performance relative to the baselines. However, we do see a consistent drop in performance as we lower $\tau$ and merge more nodes together. This is more pronounced in the larger datasets (ChEMBL, Opendata, TPC), which have more columns and evaluation pairs that can benefit from the value-cooccurrence signal. We will update the appendix with this ablation study.

\subsection{Impact of Residual Connection}

\begin{table}[h]
	\centering
	\caption{Impact of the residual connection on MRR and Recall@GT.}
	\label{tab:residual_connection}
	{
		\footnotesize
		\begin{tabular}{llrrrrrr}
			\toprule
			\textbf{Metric} & \textbf{Method} & \textbf{ChEMBL} & \textbf{Magellan} & \textbf{Opendata} & \textbf{TPC} & \textbf{Wikidata} & \textbf{SOTAB} \\
			\midrule
			\multirow{2}{*}{MRR}
			                & SemStruct       & 0.9988          & 1.0000            & 0.9995            & 0.9982       & 0.9458            & 0.6550         \\
			                & No-Res          & 0.9955          & 1.0000            & 0.9987            & 0.9977       & 0.9168            & 0.4163         \\
			\midrule
			\multirow{2}{*}{Recall@GT}
			                & SemStruct       & 0.9909          & 1.0000            & 0.9960            & 0.9968       & 0.9458            & 0.1561         \\
			                & No-Res          & 0.9863          & 1.0000            & 0.9901            & 0.9954       & 0.9146            & 0.1560         \\
			\bottomrule
		\end{tabular}
	}
\end{table}

We also examine the impact of using a residual connection between the GNN output and the initial PLM embeddings.~\Cref{tab:residual_connection} shows that the residual connection provides a consistent boost in both MRR and Recall@GT across all datasets. The improvement is particularly pronounced in the SOTAB-SM dataset, which contains more complex and noisy tables. This suggests that the residual connection helps preserve the semantic priors of the PLM while allowing the GNN to capture structural interactions, leading to better generalization in challenging scenarios.

\subsection{Runtime Scaling of Graph Construction}

\begin{table}
	\centering
	\caption{Scaling over \textit{Number of rows}}
	\label{tab:scaling_rows}
	{
		\footnotesize
		\begin{tabular}{r rr rr r}
			\toprule
			\textbf{\# rows} & \multicolumn{2}{c}{\textbf{Graph construction}} & \multicolumn{2}{c}{\textbf{GNN inference}} & \textbf{\# value nodes}                        \\
			\cmidrule(lr){2-3} \cmidrule(lr){4-5}
			                        & \textbf{Time}                                   & \textbf{Memory}                            & \textbf{Time}           & \textbf{Memory} &    \\
			\midrule
			100                     & 259.2$\pm$203.0 ms                              & 29.7 MB                                    & 7.6$\pm$2.3 ms          & 0.2 MB          & 10 \\
			500                     & 244.4$\pm$179.0 ms                              & 7.4 MB                                     & 63.0$\pm$54.0 ms        & 0.6 MB          & 10 \\
			1,000                   & 171.0$\pm$15.8 ms                               & 9.7 MB                                     & 72.3$\pm$61.0 ms        & 1.1 MB          & 10 \\
			5,000                   & 477.1$\pm$106.9 ms                              & 20.0 MB                                    & 378.5$\pm$107.0 ms      & 5.3 MB          & 10 \\
			10,000                  & 1.47$\pm$0.18 s                                 & 30.8 MB                                    & 465.4$\pm$76.3 ms       & 18.7 MB         & 10 \\
			50,000                  & 3.06$\pm$0.33 s                                 & 70.2 MB                                    & 2.04$\pm$0.10 s         & 23.1 MB         & 10 \\
			\bottomrule
		\end{tabular}
	}
\end{table}

\begin{table}
	\centering
	\caption{Scaling over \textit{Number of columns}}
	\label{tab:scaling_cols}
  {\footnotesize
	\begin{tabular}{r rr rr r}
		\toprule
		\textbf{\# columns} & \multicolumn{2}{c}{\textbf{Graph construction}} & \multicolumn{2}{c}{\textbf{GNN inference}} & \textbf{\# value nodes}                         \\
		\cmidrule(lr){2-3} \cmidrule(lr){4-5}
		                           & \textbf{Time}                                   & \textbf{Memory}                            & \textbf{Time}           & \textbf{Memory} &     \\
		\midrule
		5                          & 65.7$\pm$61.2 ms                                & ---                                        & 40.1$\pm$31.7 ms        & 0.0 MB          & 5   \\
		10                         & 88.2$\pm$35.7 ms                                & ---                                        & 60.4$\pm$43.2 ms        & ---             & 10  \\
		20                         & 408.5$\pm$267.6 ms                              & 0.5 MB                                     & 53.5$\pm$17.8 ms        & 1.8 MB          & 20  \\
		50                         & 685.9$\pm$252.8 ms                              & 34.7 MB                                    & 290.4$\pm$107.7 ms      & 0.2 MB          & 50  \\
		100                        & 1.41$\pm$0.21 s                                 & 57.3 MB                                    & 461.7$\pm$107.3 ms      & ---             & 100 \\
		200                        & 3.55$\pm$0.71 s                                 & 104.4 MB                                   & 713.0$\pm$112.4 ms      & 0.7 MB          & 200 \\
		\bottomrule
	\end{tabular}
  }
\end{table}

\begin{table}
	\centering
	\caption{Scaling over \textit{Unique values / col}}
	\label{tab:scaling_unique_values}
  {\footnotesize
	\begin{tabular}{r rr rr r}
		\toprule
		\textbf{\# Unique} & \multicolumn{2}{c}{\textbf{Graph construction}} & \multicolumn{2}{c}{\textbf{GNN inference}} & \textbf{\# value nodes}                        \\
		\cmidrule(lr){2-3} \cmidrule(lr){4-5}
		                             & \textbf{Time}                                   & \textbf{Memory}                            & \textbf{Time}           & \textbf{Memory} &    \\
		\midrule
		5                            & 22.4$\pm$1.1 ms                                 & 0.0 MB                                     & 85.3$\pm$76.6 ms        & 0.0 MB          & 10 \\
		10                           & 90.0$\pm$84.6 ms                                & 0.0 MB                                     & 43.5$\pm$17.1 ms        & 0.0 MB          & 10 \\
		50                           & 435.3$\pm$567.2 ms                              & 22.0 MB                                    & 46.4$\pm$17.5 ms        & 0.0 MB          & 10 \\
		100                          & 668.5$\pm$889.1 ms                              & 28.0 MB                                    & 41.6$\pm$14.5 ms        & 0.0 MB          & 10 \\
		500                          & 4.10$\pm$4.64 s                                 & 219.0 MB                                   & 94.7$\pm$43.1 ms        & 0.1 MB          & 10 \\
		1,000                        & 4.26$\pm$2.94 s                                 & 252.3 MB                                   & 35.8$\pm$17.4 ms        & 0.0 MB          & 10 \\
		\bottomrule
	\end{tabular}
  }
\end{table}

We also analyze the runtime scaling of our graph construction and GNN inference steps with respect to three key factors: number of rows, number of columns, and average number of unique values per column.~\Cref{tab:scaling_rows,tab:scaling_cols,tab:scaling_unique_values} present the results of this analysis. Each entry shows the average time and memory usage for graph construction and GNN inference across a range of values for each factor, while keeping the other factors fixed. We find that the graph construction time scales linearly with the number of rows and columns, which is expected given our node construction strategy. The GNN inference time also scales linearly with the number of columns, but is relatively insensitive to the number of rows due to our use of neighborhood sampling in GraphSAGE. The number of unique values per column has a more significant impact on both graph construction and inference time, as it directly affects the number of value nodes in the graph. However, even with a large number of unique values (up to 1,000), the total runtime remains within a few seconds, demonstrating the efficiency of our approach for real-world tables.

\end{document}